\documentclass{article}
\usepackage{arxiv}

\usepackage[utf8]{inputenc} 
\usepackage[T1]{fontenc}    
\usepackage[hidelinks,colorlinks=true,linkcolor=blue, citecolor=blue]{hyperref}       
\usepackage{url}            
\usepackage{booktabs}       
\usepackage{amsfonts}       
\usepackage{nicefrac}       
\usepackage{microtype}      
\usepackage{graphicx}
\usepackage[capposition=below]{floatrow}

\usepackage{natbib}
\setcitestyle{authoryear,open={(},close={)}}

\usepackage{url}            

\usepackage{algorithm,algcompatible,amsmath}
\usepackage{caption}
\usepackage{amsfonts}
\usepackage{amssymb}
\usepackage{calrsfs}
\usepackage{dsfont}

\usepackage{multirow}

\title{CommunityFish: \\ A Poisson-based Document Scaling With Hierarchical Clustering}

\author{
 Sami Diaf \\
  Faculty of Business, Economics and Social Sciences, Department of Socioeconomics\\
  Universität Hamburg\\
  \texttt{sami.diaf@uni-hamburg.de} \\
	}

\begin{document}

\maketitle  
\date{}            
\begin{abstract}
Document scaling has been a key component in text-as-data applications for social scientists and a major field of interest for political researchers, who aim at uncovering differences between speakers or parties with the help of different probabilistic and non-probabilistic approaches. Yet, most of these techniques are either built upon the agnostically bag-of-word hypothesis or use prior information borrowed from external sources that might embed the results with a significant bias. If the corpus has long been considered as a collection of documents, it can also be seen as a dense network of connected words whose structure could be clustered to differentiate independent groups of words, based on their co-occurrences in documents, known as \textit{communities}. This paper introduces \textit{CommunityFish} as an augmented version of \textit{Wordfish} based on a hierarchical clustering, namely the \textit{Louvain} algorithm, on the word space to yield \textit{communities} as semantic and independent n-grams emerging from the corpus and use them as an input to \textit{Wordfish} method, instead of considering the word space. This strategy emphasizes the interpretability of the results, since communities have a non-overlapping structure, hence a crucial informative power in discriminating parties or speakers, in addition to allowing a faster execution of the Poisson scaling model. Aside from yielding communities, assumed to be subtopic proxies, the application of this technique outperforms the classic \textit{Wordfish} model by highlighting historical developments in the \textit{U.S. State of the Union} addresses and was found to replicate the prevailing political stance in Germany when using the corpus of parties' legislative manifestos.

\end{abstract}

\section{Introduction}

Comparative politics has been a privileged domain of application of what is currently known as text-as-data field, featuring the use of text mining techniques and machine learning algorithms to identify patterns that differentiate document contents or track differences at the meta-data level. Scaling techniques consist mostly of unsupervised methods, whether probabilistic or non-probabilistic, aiming at extracting one or multiple dimensions to perform metadata comparisons \citep{Goet2019}, based on a set of assumptions concerning the word-level. In other terms, scaling techniques aim to uncover one or many hidden variables that artificially measure differences between speakers or parties, mostly based on word counts. 

Earlier scaling techniques used statistical learning approaches as for matrix factorization schemes \citep{lsa1990} or probabilistic model based on the Poisson distribution as for \textit{Wordfish} \citep{wordfish2008}, which ranks documents on a unidimensional scale using word occurrences in each document of the corpus. Further extensions of Poisson-based inferences considered a debate structure \citep{wordshoal2016}, pre-trained embedding models \citep{semscale2019}, word variations \citep{Vafa2020} or topic variability \citep{Diaf2022} that provided a sound ranking of documents depending on assumptions and use cases at the word or document levels.

As for \textit{Wordfish}, the inference uses word counts to learn a normally-distributed hidden dimension, assumed to be a proxy of partisanship among political parties when scaling manifestos \citep{wordfish2008}. Yet, the Poisson distribution does not always pertain, as frequent words are likely to have a normal distribution, while very rare words tend to follow a negative binomial distribution \citep{Lo2016}. Another noticeable drawback is the dynamic word usage which needs time-varying parameters for the Poisson ranking model and further techniques to ensure its stability \citep{Jentsch2020}, or to consider the structure \textit{document-topic-word} to get polarization at the topic level \citep{Diaf2022_3}.

Although the choice of scaling techniques is abundant, it does not necessarily meets the expectation of practitioners because the inference is done at the word-level, while the analysis often target documents' content in terms of groups of words that translate the interest of researchers, either as specific keywords or as topics. The word contributions to the built scale in \textit{Wordfish} is static and cannot be fully interpretable if the corpus' word usage underwent several changes between parties/speakers over time \citep{Jentsch2020} and the sign of specific words could be different from the position of documents they mostly represent, thus not in-line with experts' assessments \citep{Hjorth2015}. These shortcomings are imputable to the bag-of-word assumption and the underlying agnostic hypothesis of word independence that prevents a sound scaling of documents \citep{semscale2019}.

Advances in social network analysis demonstrated the efficiency of hierarchical clustering to uncover latent groups of users, or communities, that are homogeneous and distinguishable from other neighbors, based on their interactions. This concept could be easily extended to a corpus of documents to identify independent, semantic groups of words, in form of n-grams, that differentiate documents based on their occurrence while delivering informative signals that outperform analyses based on single-word usage. \textit{Louvain} algorithm \citep{Blondel2008} remains a popular clustering approach and was applied to the study social networks but also to get word groups that better represent the rhetoric used in the corpus \citep{cbail2016} or to study the lexical shift in the State Of The Union addresses \citep{Rule2015}. Similar clustering schemes were proposed as for \textit{Infomap} \citep{Infomap2008} which uses random walk map-equation instead of optimizing the modularity as for \textit{Louvain}, although the functioning of both methods was found to be similar \citep{Lancichinetti2009}. \cite{Traag_2019} proposed a faster and enhanced clustering algorithm known as \textit{Leiden} that was found to outperform \textit{Louvain} when applied to big networks.

This paper prolongs the idea of \textit{lexical shift} \citep{Rule2015} by identifying communities as representative groups of words with a high density, able to achieve a fast and interpretable scaling of documents upon which a Poisson ranking model could be easily built, instead of considering a plain word-count model relying on the bag-of-word hypothesis. I argue that communities offer a better polarization level when differentiating documents and metadata than a standard word-level model, in addition of efficiently speeding up the learning process by reducing the size of the document-term-matrix, whose sparsity may hinder the convergence of Poisson models. Regular words used by all parties/speakers are likely to form communities featuring a high number of words but less likely polarized compared to communities whose usage is exclusive, denoting the focus of a given party/author to a specific subject of item that could be uncovered without the need to run topic models. 

Hence, the proposed two-step scaling technique, \textit{CommunityFish}, brings two major novelties in the document scaling field. First, it successfully identifies communities as powerful word representations without the need to use agnostic topic models, neither running transfer learning, in terms of word embedding \citep{word2vec2013} nor adopting priors as for keywords identification \citep{keyatm2020}, but by simply running a hierarchical clustering, whether with \textit{Louvain} or \textit{Leiden} algorithms. Second, it yields a fast and interpretable scaling where communities could be easily associated with metadata without requiring much efforts as for classic word-based techniques.     

Two historical corpora, in English and German, were chosen to test this novel approach. Application on the U.S. State Of The Union (SOTU) addresses (1854-2019) shows a clear dominance of historical developments as for economic issues, local affairs and foreign policy that ranked addresses on a two-regime scale whose transition could be identified during the great depression. On the manifestos of German political parties (2013-2022), \textit{CommunityFish} uncovered granular themes at the center of election debates that were found to replicate the ideological spectrum of political parties with AFD and Linke parties being the ideological bounds of the learned scale, while other parties seem to share many featured themes, hence reinforcing their centric positions.

The paper expands the concepts used by \textit{CommunityFish} from a network analysis perspective (Section 2) and from statistical learning (Section 3), then applies the proposed algorithm to two corpora (Section 4) and compares to the standard \textit{Wordfish} used by practitioners.



\section{Network Analysis}

Analysis of social media, especially the interactions between users, drove the attention of scientists on the necessity to adopt advanced clustering methods able to extract information that describe hidden relationships between speakers via the types of messages or ideas they produce \citep{White2008}, instead of simple link structures between individuals \citep{cbail2016}. 

Network analysis witnessed important contributions on identifying distinct subgroups in social networks and several optimization schemes were developed to offer intuitive clustering \citep{Lancichinetti2009} and meaningful ranking of users.

For such tasks, researchers need to pay attention to the selected clustering methods for community detection as well as considering centrality scores \citep{Mester2021}. For large networks, \textit{Louvain} algorithm \citep{Blondel2008} is usually preferred to \textit{FastGreedy} algorithm \citep{Clauset2004} due to its relative low complexity, as it achieves a local optimization of the modularity $Q$ at the node-level, defined as : 

\centerline{
$ Q={\frac {1}{2m}}\sum \limits _{ij}{\bigg [}A_{ij}-{\frac {k_{i}k_{j}}{2m}}{\bigg ]}\delta (c_{i},c_{j})$}

with $A_{ij}$ representing the edge weight between nodes $i$ and $j$, $k_{i}$ and $k_{j}$ are the sum of the weights of the edges attached to nodes $i$ and $j$, respectively; $m$ is the sum of all of the edge weights in the graph; $c_{i}$ and $c_{j}$ are the communities of the nodes; and $\delta$ is Kronecker delta function $\delta (x,y)=1$ if x=y, 0 otherwise.

\textit{Louvain} algorithm iteratively optimizes the modularity $Q$ by starting with different node being its own community, and the concept is to place a node $n_{i}$ to one of its neighboring nodes community, in a way to maximize the modularity change \citep{Mester2021}. Similar to users in social networks, \textit{Louvain} algorithm could be applied to cluster words in a given corpus, so to extract groups of words, in a form of n-grams of different lengths, having an independent, non-overlapping structure stemming from the specific word usage found in documents.

\citet{Traag_2019} proposed \textit{Leiden} clustering as a reliable alternative, outperforming Louvain when it comes to discern small connected communities in large network structures. Altough \textit{Leiden} was found to be faster, in terms of execution, then \textit{Louvain}, both do not differ when the network structure is relatively small, as for collection of documents with limited vocabulary, meaning the community structures of both algorithms share many similarities.

\section{Poisson ranking model}

To apply \textit{CommunityFish}, the corpus is decomposed into bigrams and a minimum threshold $\pi$ is set before running \textit{Louvain} algorithm that yields $K$ communities used as features of the Document-Term-Matrix (DTM), instead of considering all words in the corpus, hence communities serve as features to the \textit{Wordfish} scaling algorithm. This scheme could be seen as a \textit{semantic clustering} of the DTM that identifies correlated pairs of words in local contexts thanks to hierarchical clustering on bigrams, which differs from a simple bigram grouping of the initial DTM features. 

The resulting DTM is given as an input to \textit{Wordfish} \citep{wordfish2008} to learn document positions, or ideal points, that scale documents based on the occurrence of communities. As a scaling technique, \textit{Wordfish} uncovers a latent scale $\theta$, assumed to be a proxy of partisanship or ideological differences between parties or speakers, depending on the used context.

While the use of the Poisson distribution is justified by the occurrence of words in the corpus, assumed to be rare events, it does not always pertain to cases where the word usage concerns few documents, meaning the Poisson's expectation departs significantly from the variance,  as demonstrated by \citet{Lo2016} who suggested the use of the Negative Binomial distribution  with resampling inference.

I argue that the use of communities frees the DTM from potential biases raised by rare words and allows a fast convergence of \textit{Wordfish} algorithm when applied to big corpora. \textit{CommunityFish} could be seen as a double dimensionality reduction technique: first to uncover communities, as the primary unit of analysis, and second to learn a scale of ideal points using \textit{Wordfish}.

\begin{algorithm}
	\caption{\textbf{:} CommunityFish}
	\begin{algorithmic}
		\STATE \textbf{1. Community detection} Run the \textit{Louvain} algorithm over the bigram features of the corpus and extract $K$ groups of words or \textit{communities}.
		\STATE \textbf{2. Wordfish} The $K$ communities are used as features for the Document-Term-Matrix, to be given as input to a Poisson Scaling Model \citep{wordfish2008} to uncover the scale $\theta_{i}$ from the specification: \\
		$log(\lambda_{ij})=\alpha_{i}+\psi_{j}+\theta_{i}\beta_{j}$, where: \\
		$\lambda_{ij}$: frequency of the community j in document i \\
		$\alpha_{i}$: document fixed effect \\
		$\psi_{j}$: community fixed effect \\
		$\theta_{i}$: the \textit{position} of document i \\
		$\beta_{j}$: the effect of community j to the document position \\
	\end{algorithmic}
\end{algorithm}

The hierarchical clustering applied to the corpus (\textit{Louvain} algorithm) could be interpreted as an implicit factorization of the traditional unigram Document-Term-Matrix (DTM) to a more interpretable feature matrix stemming from the learned communities. Aside from lowering the DTM dimension, it permits to intuitively concentrate the scaling on meaningful groups of words (communities) that discriminate the ideal points based on their occurrences in the documents, offering a more balanced basis to discern documents of the corpus based on similarities in the communities' probabilistic distributions.    

\section{Application}

Two popular datasets, with different specifications, are chosen to perform \textit{CommunityFish}, one featuring dynamic word usage as documents span over 163 years (State Of The Union) targeting several aspects of American politics, and another gathering political manifestos of main German political parties during the last three legislative elections that help identify similarities and differences in the used rhetoric in the pre-election political debates. For both corpora, the hierarchical clustering was applied on lemmatized bigrams who occur more than 30 times in the corpus, to concentrate the information on a reasonable number of communities.

\subsection{State Of The Union}

State Of The Union (SOTU) addresses consist of speeches given during the period (1854-2019), so to highlight the duality democratic-republican in the scaling \citep{Diaf2022}. The corpus was lemmatized using \textit{udpipe} model \citep{udpipe2016} to reduce the size of the Document-Term-Matrix and learn robust communities, in comparison with the raw corpus. The application of the \textit{Louvain} algorithm yielded 52 different communities\footnote{\textit{Leiden} clustering yielded a similar community structure to \textit{Louvain}, with minor differences concerning two communities, out of 52. The same results was found using the German political manifesto corpus.} (Table 1) with a historical context that spans over one and half century, with different episodes of modern American history. Noticeable is that 22, out of 52, communities are constituted of bigrams and the remaining are n-grams of different lengths. 

Communities, whose contributions to the scale $\beta_{j}$ are different from zero, polarize the overall scale $\theta$ via their respective signs. From Figure 1, communities 45, 40, 11 and 8 contribute to documents whose positions in the overall scale (Figure 2) are positive, consisting of earlier addresses from the second half of the Ninetieth century targeting neighboring countries and local administration, while modern addresses have negative positions (Figure 2) and demonstrate a strong influence of foreign policy and defense interests (communities 38 and 49) as well as business/economic environment (communities 43 and 2). Figure 2 shows a two-regime scale, whose transition could be identified during the great depression (Hoover addresses during the period 1929-1933, coinciding with the position $\hat{\theta} =0$), indicating a potential shift in the rhetoric, or transition into modern addresses, used by US presidents and captured via communities that represent proxies of most tackled interests during the addresses. The classic Wordfish (Figure 5) yields more clustered document positions that cannot not be differentiated in small periods, even if given by different speakers.

\begin{table}
	\begin{tabular}{ cc }   
		Table1: Communities in SOTU corpus & Table 2: Communities in German Manifesto corpus \\  
		\scalebox{0.60}{
			\begin{tabular}{|l|l|}
				\hline
				\textbf{Community} & \textbf{Words} \\
				\hline
				com\_1 & agricultural, product\\
				\hline
				\multirow{ 2}{*}{com\_2}  & american , billion  , business , enlist   , every    , fellow   , million  , silver   , small  \\ &  young    , citizen  , family   , people   , republics, dollar   , man      , day      , americans\\
				\hline
				com\_3 & annual , special, message\\
				\hline
				com\_4 & armed   , military, naval   , force\\
				\hline
				\multirow{ 2}{*}{com\_5} & ask     , come    , current , end     , fiscal  , five    , four    , last    , many    , next    , past    precede ,\\ & previous, recent  , ten     , three   , two     , year    , congress, june    , session , ago     , ahead\\
				\hline
				\multirow{ 3}{*}{com\_6} & attorney  , british   , can       , federal   , general   , government, local     , make      ,  must    \\ &  national  , postmaster, self      , social    , spanish   , supreme   , help      , court     , sure    \\ &   also     , continue  , bank      , defense   , security\\
				\hline
				com\_7 & balanced, budget\\
				\hline
				com\_8 & base     , call     , confer   , depend   , enter    , impose   , urge     , upon     , attention\\
				\hline
				com\_9 & careful      , favorable    , consideration\\
				\hline
				com\_10 & central, latin  , south  , america\\
				\hline
				\multirow{ 2}{*}{com\_11} & civil    , hard     , human    , interest , postal   , public   , right    , tax      , work     , service  , rate \\ &  debt     , building , land     , opinion  , now      , credit   , cut      , reduction, together\\
				\hline
				com\_12 & commerce  , interstate, commission\\
				\hline
				com\_13 & earnestly, recommend\\
				\hline
				com\_14 & economic   , development, growth\\
				\hline
				com\_15 & executive, branch   , order\\
				\hline
				com\_16 & exist        , international, law          , present      , tariff       , enforcement  , condition    , system\\
				\hline
				com\_17 & far  , thus , reach\\
				\hline
				com\_18 & first, time\\
				\hline
				\multirow{ 2}{*}{com\_19} & foreign   , free      , great     , nation    , office    , post      , take      , treasury  , war       , world    \\ & country  com\_    , power     , trade     , britain   , department, place     , ii\\
				\hline
				com\_20 & full      , employment\\
				\hline
				com\_21 & go     , look   , move   , forward\\
				\hline
				com\_22 & god  , bless\\
				\hline
				com\_23 & good , faith\\
				\hline
				com\_24 & health   , medical  , care     , insurance\\
				\hline
				com\_25 & high    , level   , priority, school\\
				\hline
				com\_26 & internal, revenue\\
				\hline
				com\_27 & large , number, part\\
				\hline
				com\_28 & let, us\\
				\hline
				com\_29 & long, run , term\\
				\hline
				com\_30 & low   , income\\
				\hline
				com\_31 & may , well\\
				\hline
				com\_32 & merchant, marine\\
				\hline
				com\_33 & middle, class , east\\
				\hline
				com\_34 & minimum, wage   , worker\\
				\hline
				com\_35 & mr     , speaker\\
				\hline
				com\_36 & natural , resource\\
				\hline
				com\_37 & new    , job    , program, york\\
				\hline
				com\_38 & nuclear, weapon\\
				\hline
				com\_39 & one    , half   , hundred, third\\
				\hline
				com\_40 & panama, canal\\
				\hline
				com\_41 & per  , annum, cent\\
				\hline
				com\_42 & philippine, islands\\
				\hline
				com\_43 & private   , enterprise, sector\\
				\hline
				com\_44 & progress, step    , toward\\
				\hline
				com\_45 & puerto, rico\\
				\hline
				com\_46 & set  , forth\\
				\hline
				com\_47 & several, united , states , nations\\
				\hline
				com\_48 & sink, fund\\
				\hline
				com\_49 & soviet, union\\
				\hline
				com\_50 & vice     , president\\
				\hline
				com\_51 & welfare, reform\\
				\hline
				com\_52 & white, house\\
				\hline
		\end{tabular}} &  
		\scalebox{0.60}{
			\begin{tabular}{|l|l|}
				\hline
				\textbf{Community} & \textbf{Words} \\
				\hline
				com\_1 & abkomme , abkommen\\
				\hline
				com\_2 & afd     , demokrat, deshalb , fordern , frei    , linke   , stehen  , setzen\\
				\hline
				\multirow{2}{*}{com\_3} & alt        , brauchen   , immer      , jung       , mehr      , mensch \\&     , million    , gerechen   , stark      , geld       , personal   , transparenz, zeit\\
				\hline
				\multirow{2}{*}{com\_4} & arbeit          , beruflich       , gut             , kulturell       , selbstbestimmt  , arbeiten \\ &       , bildung         , arbeitsbedingung, leben           , zukunft\\
				\hline
				com\_5 & arbeitgeber , arbeitnehmer, patient     , verbraucher , innen\\
				\hline
				\multirow{2}{*}{com\_6} & arbeitsplatz , dass , deutschland , einsetzen    , ganz , gestalten \\ & , jed , neu , schaffen , sicherstellen,  sorgen  verhindern \\ &, zeigen , einzeln , form  , kind , technologie\\
				\hline
				\multirow{2}{*}{com\_7} & beitrag         , bund            , dabei           , etwa            , gelten          , gerade          , gesellschaftlich, \\ & insbesondere    , land            , mittel         , projekt         , regelung    \\ &     , sollen          , sowie           , teilhabe        , wichtig         , zugang          , leisten         , na              , mitteln         , rolle\\
				\hline
				com\_8 & bezahlbar, wohnraum\\
				\hline
				com\_9 & biologisch, vielfalt\\
				\hline
				com\_10 & cdu, csu\\
				\hline
				com\_11 & corona, krise\\
				\hline
				com\_12 & demokratisch, kontrolle\\
				\hline
				com\_13 & deutsch  , bundestag, sprache\\
				\hline
				\multirow{2}{*}{com\_14} & digital         , it              , sozial          , infrastruktur   , welt            , sicherheit      , absicherung  \\ &    ,  gerechtigkeit  , marktwirtschaft , netzwerk        , sicherungssystem \\ &, wohnungsbau     , zusammenhalt\\
				\hline
				com\_15 & drei     , euro     , letzt    , milliarde, mrd      , pro      , seit     , vergangen, vier     , zehn     , jahr\\
				\hline
				com\_16 & erhalten, bleiben\\
				\hline
				com\_17 & erneuerbare , erneuerbaren, energie     , energien\\
				\hline
				com\_18 & erst   , schritt\\
				\hline
				com\_19 & eu           , ebene        , kommission   , mitgliedstaat, staat\\
				\hline
				com\_20 & fair      , wettbewerb\\
				\hline
				com\_21 & gering     , hoch       , mittler    , einkommen  , unternehmen\\
				\hline
				com\_22 & gesetzlich        , mindestlohn       , rent              , rentenversicherung\\
				\hline
				com\_23 & gleich, recht , chance, lohn  , rechte\\
				\hline
				com\_24 & hartz, iv\\
				\hline
				com\_25 & lage     , versetzen\\
				\hline
				com\_26 & medizinisch, versorgung\\
				\hline
				com\_27 & nachhaltig    , wirtschaftlich, entwicklung\\
				\hline
				com\_28 & offen       , gesellschaft\\
				\hline
				com\_29 & qualitativ, hochwertig\\
				\hline
				com\_30 & rechnung, tragen\\
				\hline
				com\_31 & rechtlich, rundfunk\\
				\hline
				com\_32 & regel , regeln\\
				\hline
				com\_33 & schnell , internet\\
				\hline
				com\_34 & schon, heute\\
				\hline
				com\_35 & schwarz, gelb\\
				\hline
				com\_36 & sexuell     , orientierung\\
				\hline
				com\_37 & start, ups\\
				\hline
				com\_38 & stelle , stellen\\
				\hline
				com\_39 & strukturschwach, region\\
				\hline
				com\_40 & stunde , stunden\\
				\hline
				com\_41 & teil  , teilen\\
				\hline
				com\_42 & treffen, triefen\\
				\hline
				com\_43 & verein  , vereinen\\
				\hline
				com\_44 & vereint, nation\\
				\hline
				com\_45 & vgl    , kapitel\\
				\hline
		\end{tabular}} \\
	\end{tabular}
\end{table}

\begin{figure}
	\centering
	\hspace*{-0.4cm}\includegraphics[width=10cm, angle=-90]{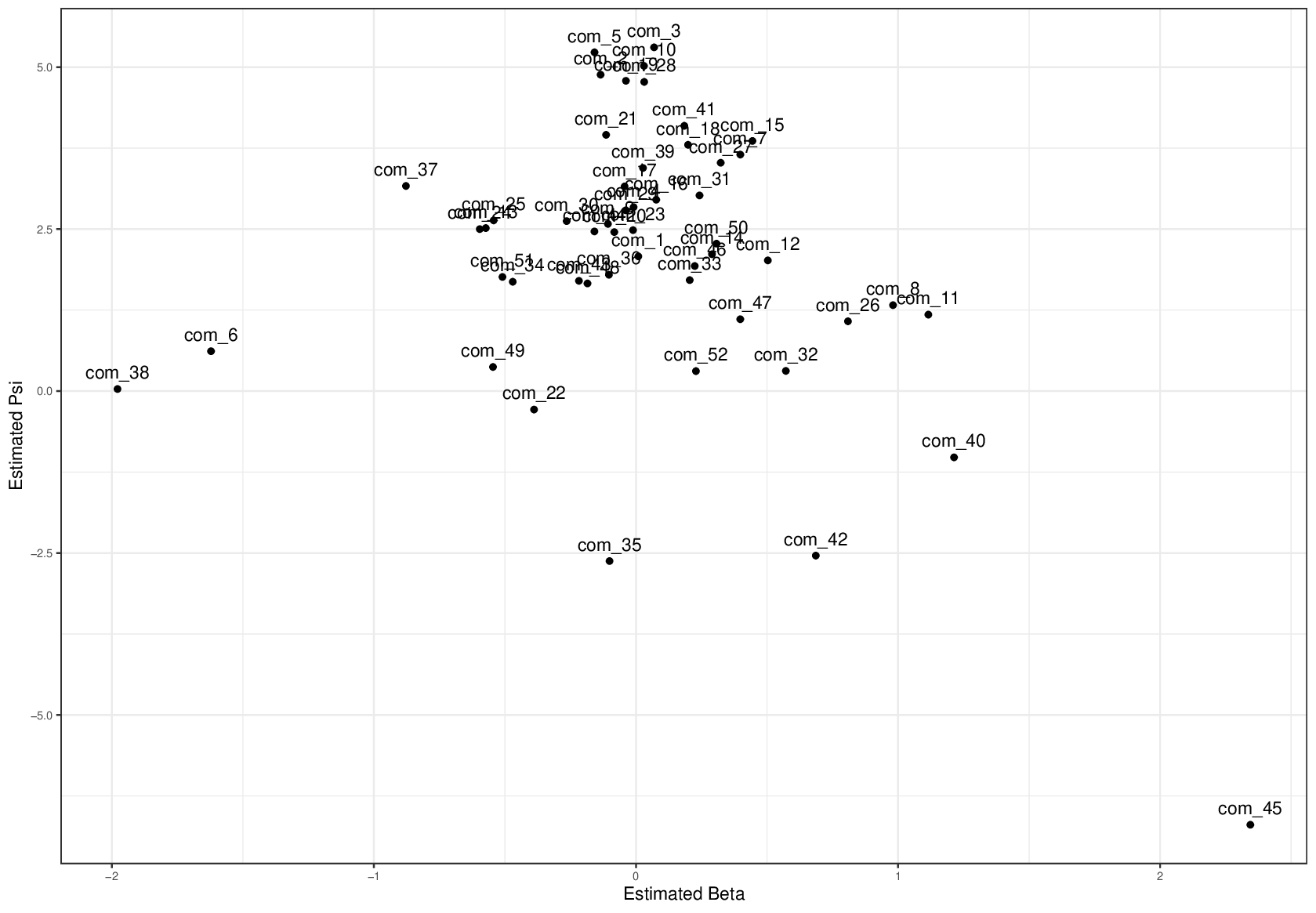}
	\caption{Communities' contributions to the scale ($\beta$) vs communities' positions $\psi$ (SOTU corpus)}\label{fig1}
\end{figure}

\begin{figure}
	\centering
	\hspace*{-0.4cm}\includegraphics[width=11cm, angle=-90]{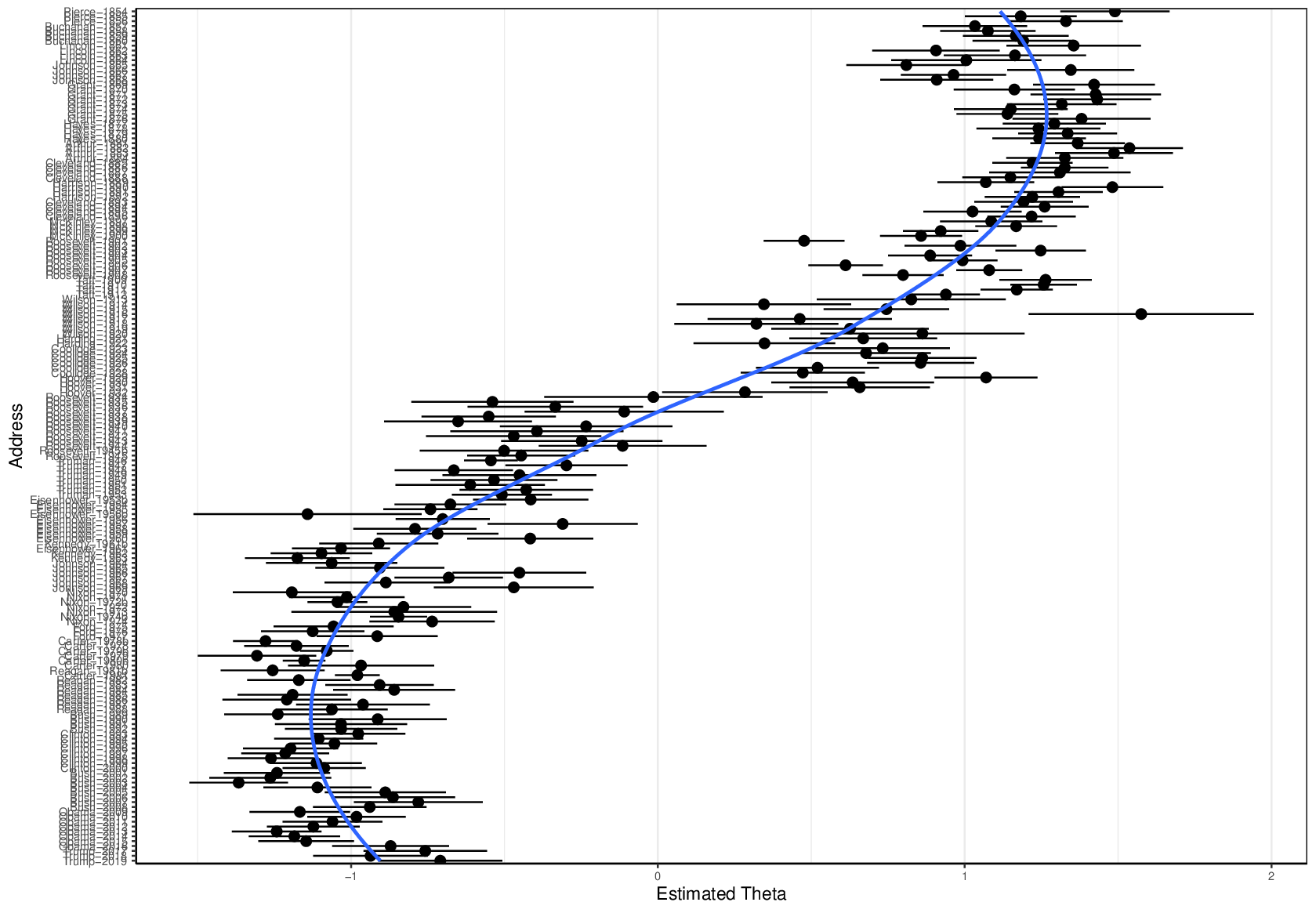}
	\caption{Learned \textit{CommunityFish} ideal points with 95\% confidence intervals (SOTU Corpus).}\label{fig2}
\end{figure}

\subsection{German Manifesto}

The corpus of Manifesto Project \citep{Manifesto2021} was used to get the manifestos of the six main German political parties, during the period 2013-2021 \citep{Diaf2022_3}, then lemmatized using a German language model \citep{udpipe2016} to reduce the vocabulary length of the corpus. It resulted 45 communities (Table 2) reproducing most of the debated themes in social life, politics and economic development which constitute the basis of the learned scale (Figure 4), found to replicate the prevailing political partisanship in Germany. AFD and Linke parties are the two bounds of the learned scale, while other parties have central positions, with noticeable firm positions (small standard deviations of their ideal points) of the Linke and Gruene parties throughout the studies period, while AFD and CDU positions seem to have the highest variability (wide standard errors). The blue line in Figure 4 is the Loess curve used to separate parties into two distinct classes (left-right) based on learned scale from the established communities (Table 2), resulting into a bi-partisanship AFD-CDU-FDP and SPD-Gruene-Linke. As a comparison to Wordfish (Figure 6), CommunityFish highlights the polarization AFD-Linke better than Wordfish, whose scale does not demonstrate a clear partisanship.

From Figure 2, communities 40 and 45 support the position of the Linke party, as their contribution to the scale is strongly positive, in comparison to communities 5, 11 and 12 whose $\beta_{j}$ are still positive but rather close to the origin. Most of the learned communities have a low contribution to the scale ($\beta_{j}$ $\rightarrow$ 0) and denote shared interests of political parties.

\begin{figure}
	\centering
	\hspace*{-0.4cm}\includegraphics[width=10cm, angle=-90]{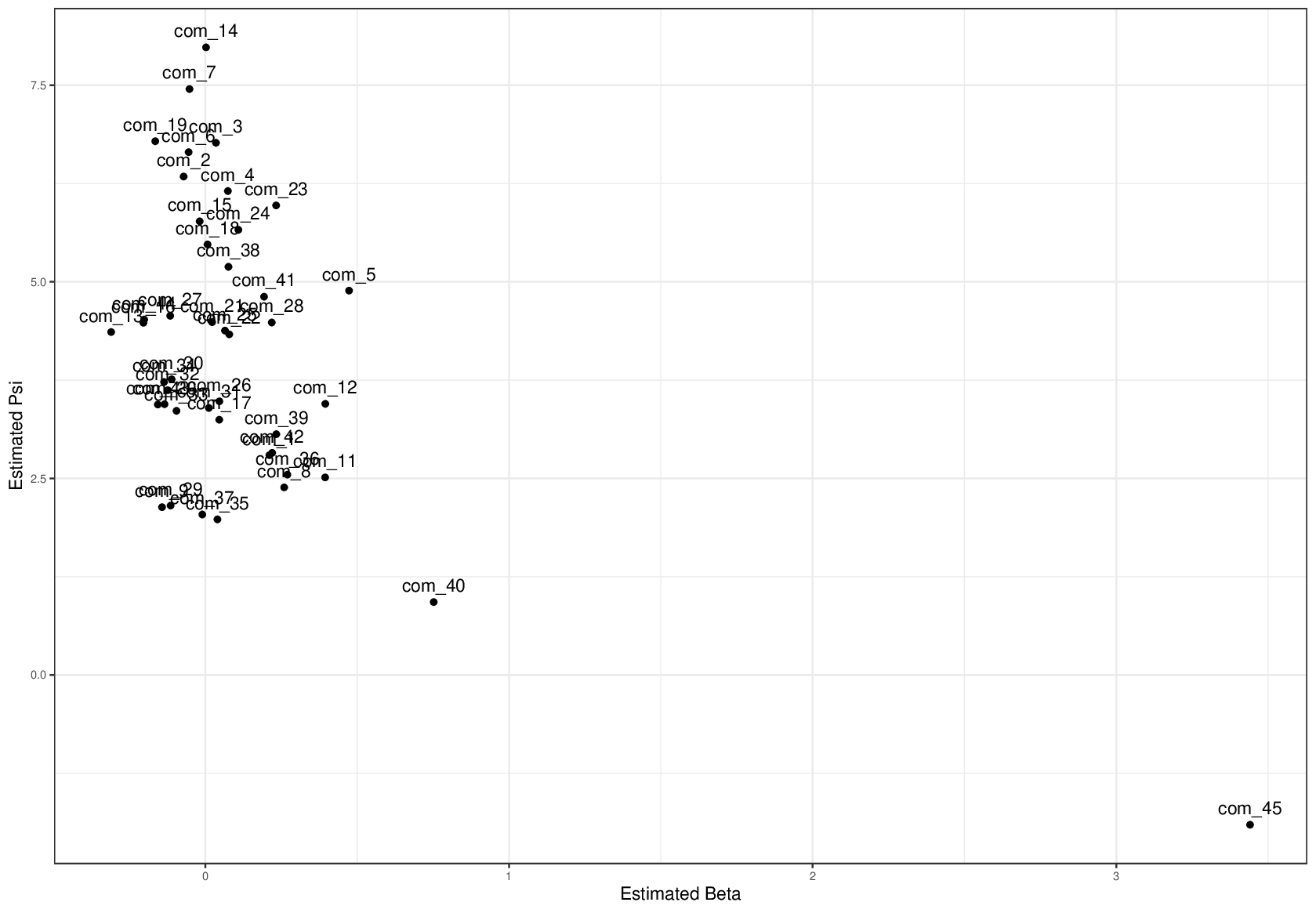}
	\caption{Communities' contributions to the scale ($\beta$) vs communities' positions $\psi$ (German Manifesto corpus)}\label{fig3}
\end{figure}

\begin{figure}
	\centering
	\hspace*{-0.4cm}\includegraphics[width=9cm, angle=-90]{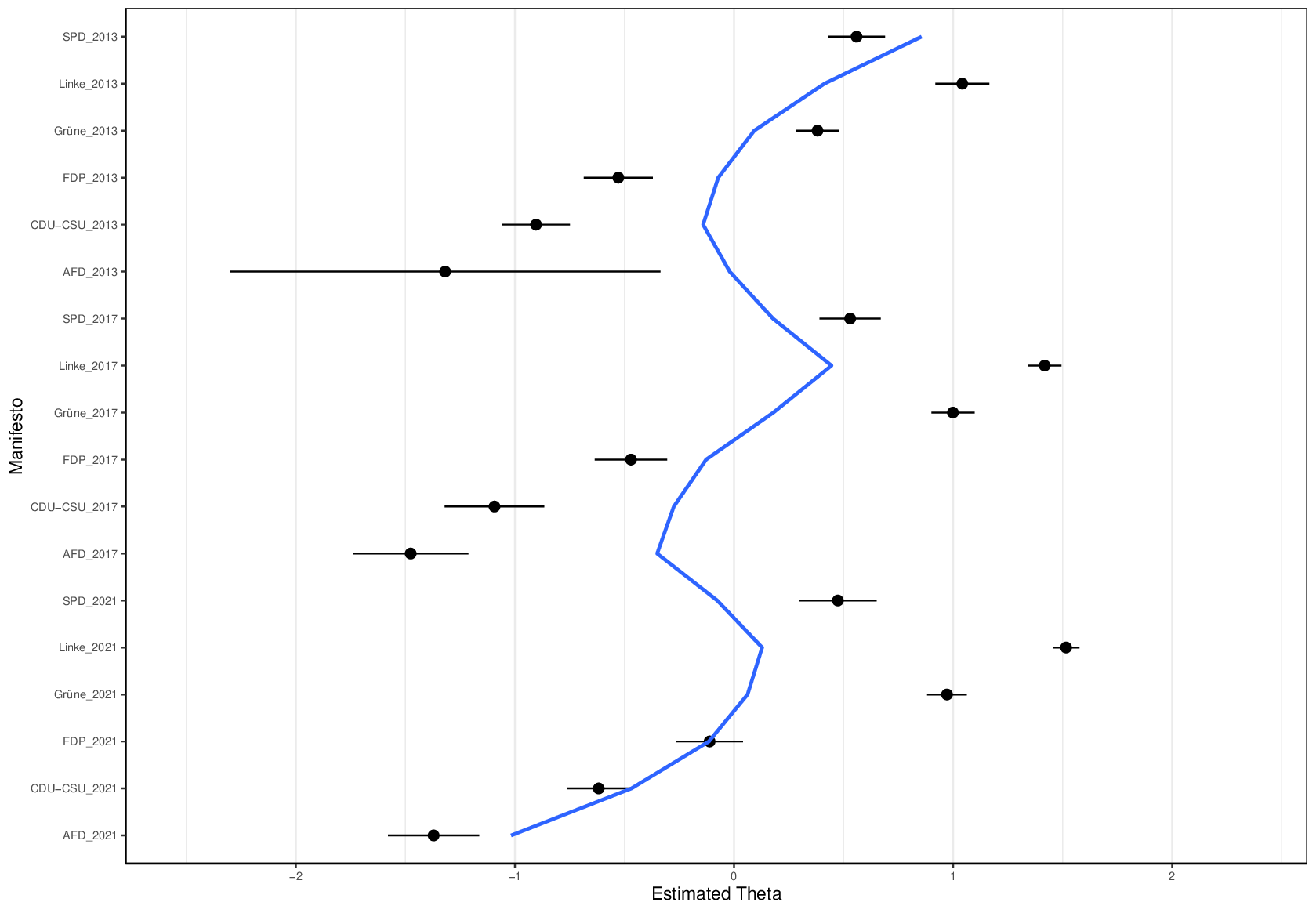}
	\caption{Learned CommunityFish ideal points with 95\% confidence intervals (German Manifesto Corpus). Blue line is the Loess curve.}\label{fig4}
\end{figure}

\begin{figure}
	\centering
	\hspace*{-0.4cm}\includegraphics[width=9cm, angle=-90]{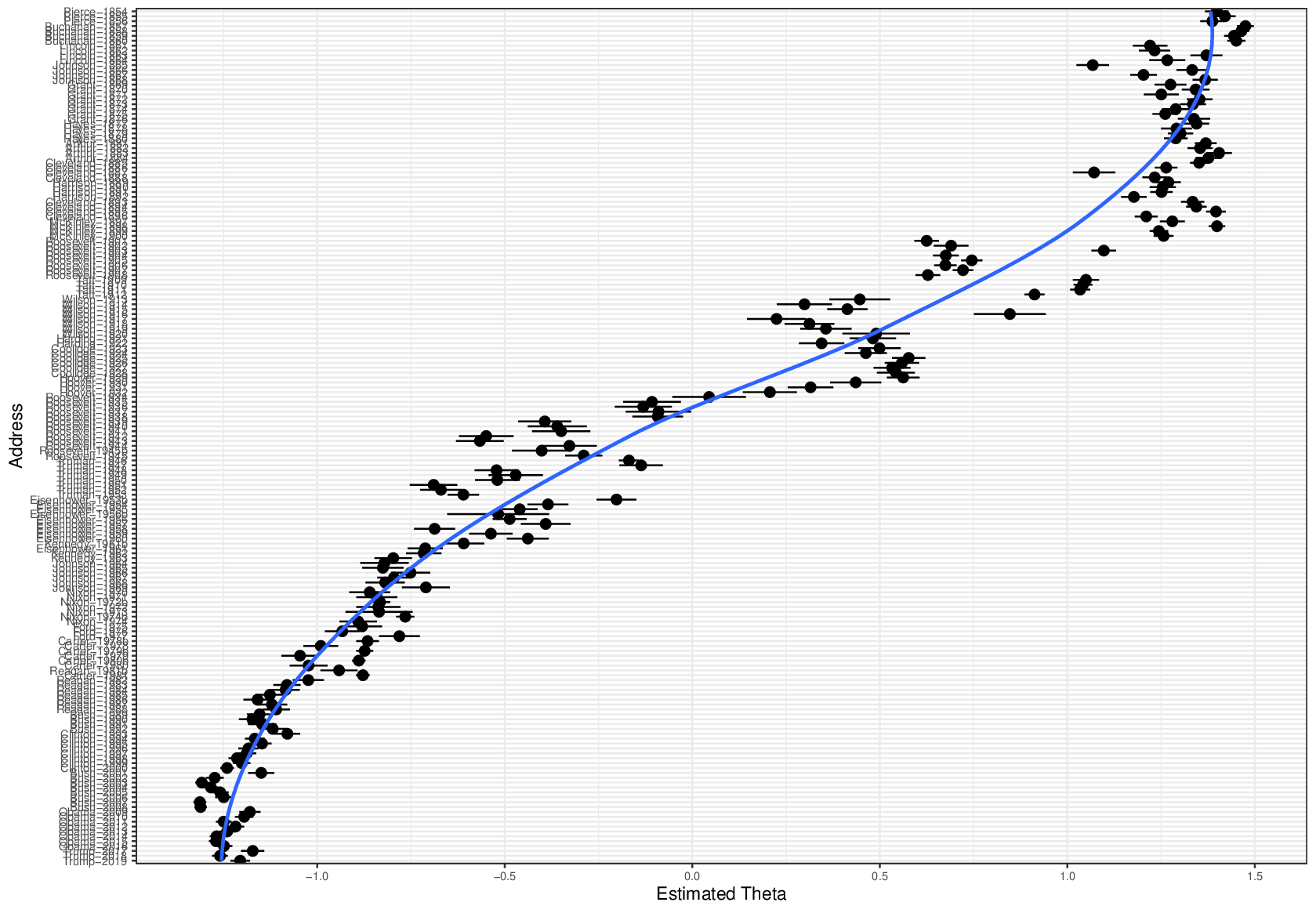}
	\caption{Learned Wordfish ideal points with 95\% confidence intervals (SOTU Corpus). Blue line is the Loess curve.}\label{fig5}
\end{figure}

\begin{figure}
	\centering
	\hspace*{-0.4cm}\includegraphics[width=9cm, angle=-90]{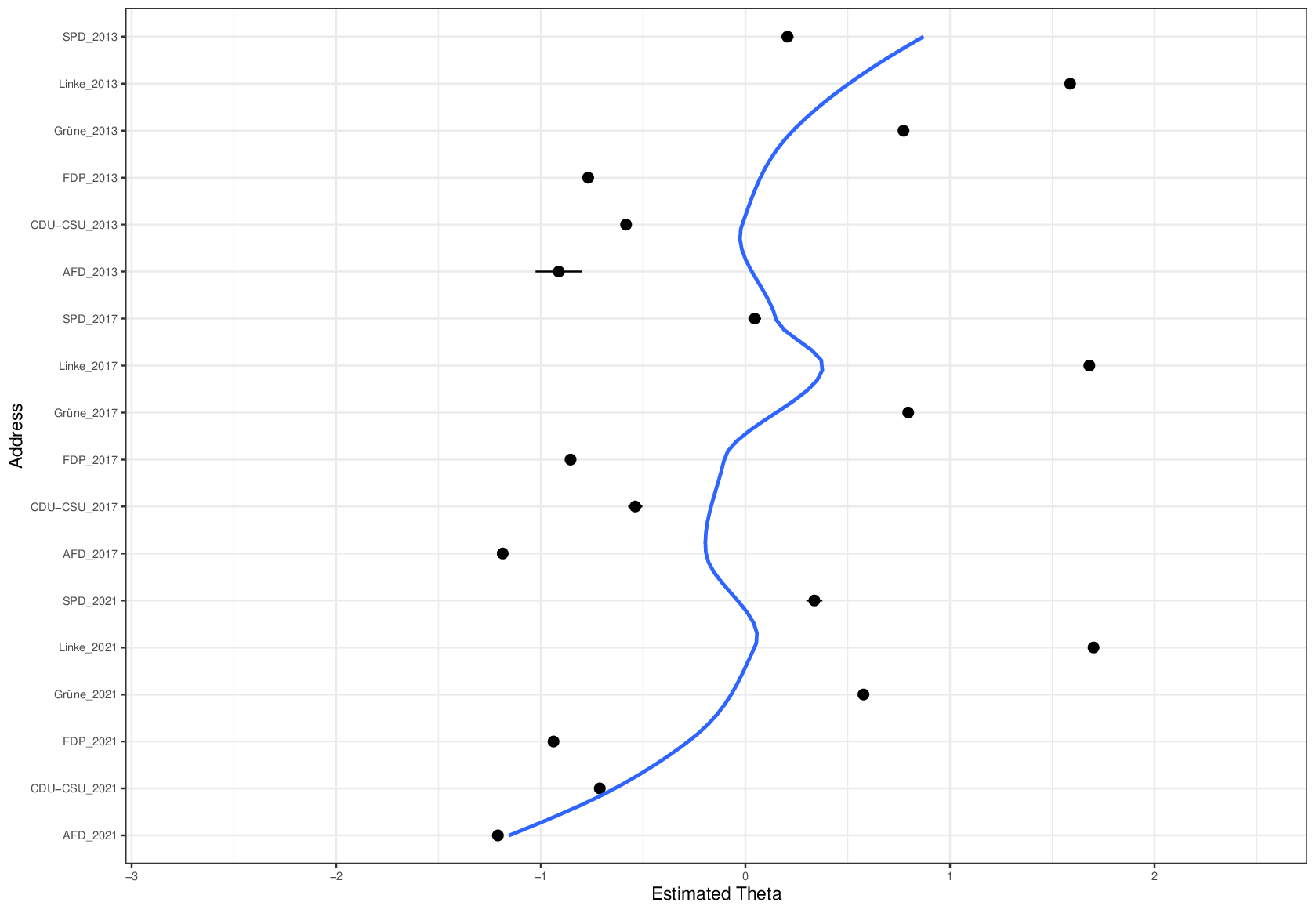}
	\caption{Learned \textit{Wordfish} ideal points with 95\% confidence intervals (German Manifesto Corpus). Blue line is the Loess curve.}\label{fig6}
\end{figure}

%

%

\clearpage

\section{Conclusion}

Scaling techniques remain useful application tools used by political scientists to investigate partisanship among parties and to study the ideological spectrum of speakers, but still suffering from the limitation of considering words as the sole unit of analysis. While many solutions were devised to enhance scaling results using external information from borrowed corpora, the use of hierarchical clustering, as a pre-processing step, permits to identify \textit{communities}, as robust groups of associated words, that proved semantically efficient in delivering substantial and interpretable results, coupled with a faster execution time. \textit{CommunityFish} is a summarization scaling technique that translates the unit of analysis from words to communities, also regarded as an implicit factorization of the document-feature-matrix, unveiling informative sub-topic structures for an in-depth scaling of historical corpora as well as political manifestos. Optimal use of \textit{CommunityFish} requires selecting most informative communities in an already-lemmatized corpus by mean of a clustering technique (could it be \textit{Louvain} or \textit{Leiden} algorithms). This ensures an independent community structure when aggregating the document-feature-matrix, helping the spread of the ideological stance learned via Poisson rank model, so to outperform classic \textit{Wordfish} algorithm without the need to adopt advanced, and usually expensive, NLP solutions that may bring extra biases from other corpora or tasks. Applied to two different corpora, it showed a great ability in extracting communities from a language-variable corpus (SOTU) and identifying common items in debate-based documents (German manifesto) for an efficient and meaningful scaling of documents.

\bibliographystyle{unsrt} 
\bibliography{biblio}

\end{document}